\documentclass[letterpaper]{article} 
\usepackage{aaai23}  
\usepackage{times}  
\usepackage{helvet}  
\usepackage{courier}  
\usepackage[hyphens]{url}  
\usepackage{graphicx} 
\usepackage{natbib}  
\usepackage{caption} 
\usepackage{algorithm}
\usepackage{amsmath}
\usepackage{amssymb}
\usepackage{xcolor}
\usepackage{comment}
\usepackage{amsmath,amssymb} 
\usepackage{color}
\usepackage{enumitem}
\usepackage{adjustbox}
\usepackage{subcaption}
\usepackage{amsmath}
\usepackage{amssymb}
\usepackage{booktabs}
\usepackage{algorithm}
\usepackage{tabu}
\usepackage{algpseudocode}
\usepackage{amsmath}
\usepackage{amssymb}
\usepackage{booktabs}
\usepackage{graphicx}
\usepackage{multirow}
\usepackage{subcaption}
\usepackage{bbm}
\usepackage{newfloat}
\usepackage{listings}
\DeclareCaptionStyle{ruled}{labelfont=normalfont,labelsep=colon,strut=off} 
\floatstyle{ruled}
\newfloat{listing}{tb}{lst}{}
\floatname{listing}{Listing}
\title{Hyperspherical Loss-Aware Ternary Quantization}

\author {
    Dan Liu, 
    Xue Liu 
}
\affiliations {
    McGill University\\
    daniel.liu@mail.mcgill.ca, xue.liu@mcgill.ca
}

\usepackage{bibentry}

\begin{document}

\maketitle

\begin{abstract}
Most of the existing works use projection functions for ternary quantization in discrete space. Scaling factors and thresholds are used in some cases to improve the model accuracy. However, the gradients used for optimization are inaccurate and result in a notable accuracy gap between the full precision and ternary models. To get more accurate gradients, some works gradually increase the discrete portion of the full precision weights in the forward propagation pass, e.g., using temperature-based Sigmoid function. Instead of directly performing ternary quantization in discrete space, we push full precision weights close to ternary ones through regularization term 
prior to ternary quantization. In addition, inspired by the temperature-based method, we introduce a re-scaling factor to obtain more accurate gradients by simulating the derivatives of Sigmoid function. The experimental results show that our method can significantly improve the accuracy of ternary quantization in both image classification and object detection tasks.
\end{abstract}

\section{Introduction}
Most deep neural network (DNN) models have a huge amount of parameters, making it impractical to deploy them on edge devices. When deploying DNN models with latency, memory, and power restrictions, the inference efficiency and model size are the main obstacles. There are many studies on how to use quantization and pruning to reduce model size and computation footprint.

DNN model compression has always been an important area of research. For instance, quantization and pruning are frequently used because they can minimize model size and resource requirements. Pruning can retain higher accuracy at the expense of a longer inference time, while quantization can accelerate and compress the model at the expense of accuracy. The purpose of model quantization is to get a DNN model with maximum accuracy and minimum bit width. Low-bit quantization has the advantage of quick inference, but accuracy is often reduced by inaccurate gradients \cite{yin2019understanding} that are estimated using discrete weights \cite{gholami2021survey}. Compared to the low-bit method, weight sharing quantization \cite{stock2019killthebits,martinez_2020_pqf,cho2021dkm} has less model accuracy drop, while inference time is longer because the actual weight values used are remain full precision.

The discrepancy between the quantized weights in the forward pass and the full precision weights in the backward pass can be reduced by gradually discretizing the weights \cite{louizos2018relaxed,jang2016categorical,chung2016hierarchical}. However, it only works well with 4-bit (or higher) precision, as the ultra-low bit (binary or ternary) quantizer may drastically affect the weight magnitude and lead to unstable weights \cite{gholami2021survey}. Recent research indicates that the essential semantics in feature maps are preserved by direction information \cite{liu2016large,liu2017deephyperspherical,liu2017sphereface,TimRDavidson2018HypersphericalVA,deng2019arcface,SungWooPark2019SphereGA,BeidiChen2020AngularVH}. 

In this work, inspired by the hyperspherical learning \cite{liu2017deephyperspherical}, we rely on the direction information of weight values to perform ternary quantization. Before performaing quantization, we use a regularization term to reduce the cosine similarity between the full precision weight values and their ternary masks under hyperspherical settings \cite{liu2017deephyperspherical}.
Then we use the proposed gradient scaling method with the straight-through estimator (STE) \cite{bengio2013estimating} to fulfill the ternary quantizaiton.
The following is a summary of our contributions:
\begin{itemize}[topsep=0pt]

\item We propose a loss-aware ternary quantization method which uses loss regularization terms to reduce the cosine distance between full precision weight values and their ternary counterparts. Once the training is done, the weight values will be separated into three clusters, aloowing for a more accurate quantization. 

\item We propose a learned quantization threshold to improve the weight sparsity and the quantization performance. We use a re-scaling method during backpropagation to simulate the temperature-based method to obtain more accurate gradients. 
\end{itemize}
\section{Related Work}
To reduce model size and accelerate inference, a notable amount of research has been devoted to model quantization methods, such as binary quantization~\cite{courbariaux2015binaryconnect,rastegari2016xnor} and ternary quantization~\cite{hwang2014fixed,li2016twn,zhu2016ttq}.
In ternary models training, optimization is difficult because the discrete weight values hinder efficient local-search. Gradient projection method such as straight-through estimator (STE) is usually used to overcome this difficulty:
\begin{equation}
\label{eq:ste}
\left\{\begin{array}{l}\mathbf{\hat{W}_{t}}=\operatorname{Proj}\left(\mathbf{{W}_{t}}\right)=\operatorname{sign}\left(\mathbf{{W}_{t}}\right) \\ \mathbf{{W}_{t+1}}=\mathbf{{W}_{t}}-\eta_{t} \hat{\nabla} L\left(\mathbf{\hat{W}_{t}}\right)\end{array}\right..
\end{equation}
$\operatorname{Proj}$ is the projection operator and projects $\mathbf{{W}} \in \mathbb{R}$ to a discrete $\mathbf{\hat{W}} \in \{0,\pm 1\}$. The optimization of $\operatorname{Proj}$ is equivalent to:
\begin{equation}
\underset{\alpha, \mathbf{\hat{W}}}{\arg \min }\|\mathbf{W}-\alpha \mathbf{\hat{W}}\|_{2}^{2}, 
\end{equation} where $\alpha$ is the scaling factor \cite{bai2018proxquant,parikh2014proximal,li2016twn}. A fixed threshold $\Delta$ is often introduced by previous works \cite{li2016twn,zhu2016ttq} to determine the quantization intervals of $\operatorname{Proj}$ (Eq.~\eqref{eq:ste}). Therefore, ternary quantization can be divided into estimation and optimization-based methods depending on how we obtain $\alpha$ and $\Delta$. 
\subsection{Estimation-Based Ternary Quantization}
The estimation-based methods, such as \cite{li2016twn}, use the approximated form $\Delta =0.7 \times \boldsymbol{E}\left(\left|\tilde{w}_{l}\right|\right)$ and 
$\alpha =\underset{w>\Delta}{\boldsymbol{E}}(|w|)$ as optimizing $\alpha$ and $\Delta$ are time consuming. \cite{wang2018two_27} uses an alternating greedy approximation method to improve $\alpha$ and $\Delta$. Given $\Delta$ the $\alpha$ has a closed form optimal solution $\alpha =\underset{w>\Delta}{\boldsymbol{E}}(|w|)$ \cite{li2016twn,zhu2016ttq}. However, direct or alternating estimation is a very rough approximation for $\Delta$. The best optimization result of Eq.~\eqref{eq:ste} cannot be guaranteed.
\subsection{Optimization-Based Ternary Quantization}
The optimization-based method TTQ \cite{zhu2016ttq} uses a fixed $\Delta=0.05\times\operatorname{max}(|\mathbf{W}|)$ and two SGD-optimized scaling factors to improve the quantization results. The intuition is straightforward: since $\Delta$ is an unstable approximation, we can use SGD to optimize the scaling factor $\alpha$. The following works \cite{yang2019quantization_22,dbouk2020dbq_20,li2020rtn_23} add more complicated scaling factors and provide corresponding optimization schemes. 

There is another subclass in the optimization-based methods, which aims to manipulate the gradient and filter out the less salient weight values by adding regularization terms to the loss function. For example, \cite{hou2018loss_31} uses second-order information to rescale the gradient to find out the weight values that are not sensitive to the ternarization, i.e. the less salient weight values.  \cite{zhou2018explicit_26.5} adds a regularization term to refine the gradient through the L1 distance. If $w$ is close to $\hat{w}$, the regularization should be small, otherwise the regularization should be large. The intuition is simple: if $w$ and $\hat{w}$ are close to each other, they should not change frequently. However, the discrete ternary values are always involved in this kind of optimization-based methods, so the gradient is not accurate, which affects the optimization results. More advanced methods \cite{dbouk2020dbq_20,yang2019quantization_22} introduce a temperature-based Sigmoid function after ternary projection in order to gradually discretize $w$. Since part of the optimization process is in continuous space, more accurate $a$ and $\Delta$ can be obtained after the weights are fully discretized. One advantage of using the Sigmoid function is that the gradient is rescaled during backpropagation, i.e., 
as $w$ approaches $\hat{w}$, the gradient approaches 0, and as $w$ approaches $\Delta$, the gradient gets larger. However, this Sigmoid-based method is very time-consuming as it performs layer-wise training and quantization. It is infeasible for very deep neural networks.

Unlike the above methods, our proposed method optimizes the distance between $\mathbf{W}$ and $\mathbf{\hat{W}}$ in continuous space to overcome inaccurate gradients before the conventional ternary quantization. Our method has similar advantages as the Sigmoid-based methods: the use of full-precision weights to optimize the distance. Compared with their time-consuming layer-wise training strategy, our method works globally and requires much less training epoch. In addition, we propose a novel way to re-scale gradients during backpropagation.


\section{Preliminary and Notations}
\subsection{Hyperspherical Networks}
A hyperspherical neural network layer \cite{liu2017deephyperspherical} is defined as:
\begin{equation}
\label{eq:spop}
    \mathbf{y}=\phi(\mathbf{W}^\top\mathbf{x}),
\end{equation}
where $\mathbf{W} \in \mathbb{R}^{m\times{n}}$ is the weight matrix, $\mathbf{x}\in \mathbb{R}^{m}$ is the input vector to the layer, $\phi$ represents a nonlinear
activation function, and $\mathbf{y}\in \mathbb{R}^{n}$ is the output feature vector.
The input vector $\mathbf{x}$ and each column vector $\mathbf{w}_j\in\mathbb{R}^{m}$ of $\mathbf{W}$
subject to $\|\mathbf{w}_j\|_2=1$ for all $j = 1, . . . , n$, and $\|\mathbf{x}\|_2=1$.
\subsection{Ternary Quantization}
To adapt to the hyperspherical settings, the ternary quantizer is defined as:
\begin{equation}
\label{eq_ter}
\hat{\mathbf{W}}=\texttt{Ternary}(\mathbf {W},\Delta)=\left\{\begin{aligned} {\alpha} &: {w_{ij}}>~~~\Delta, \\ 0 &:\left|{w_{ij}}\right| \leq ~\Delta, \\-{\alpha} &: {w_{ij}}<-\Delta, \end{aligned}\right.
\end{equation}
\begin{equation}
    s.t. ~~ \alpha=\frac{1}{\sqrt{\|\mathbf{w}_j\|_0}},
\end{equation}
where $\Delta$ is a quantization threshold, $\|\mathbf{w}_j\|_0$ denotes the number of non-zero values in $\mathbf{w}_j$, and $\|\mathbf{\hat{w}}_j\|_2=1$. The ternary hyperspherical layer has the following property:
$   \label{hcode}
    \phi(\mathbf{\hat{w}}_j^\top\mathbf{x})=\phi(\alpha\mathbf{\bar{w}}_j^\top\mathbf{x}),
$
where $\mathbf{\bar{w}}\in {\{-1,0,1\}}$.
\section{Loss-Aware Ternary Quantization}
The magnitude and direction information of weight vectors change dramatically during ternary quantization. Hyperspherical neural networks can work without taking the magnitude change into account. Combining ternary quantization with hyperspherical learning may offer stable features beneficial for quantization performance.

In this section, to relieve the magnitude and direction deviation in the subsequent quantization procedure, we first introduce a regularization term to push full precision weights toward their ternary counterparts. Then, a re-scaling factor is applied during the quantization stage to obtain more accurate gradients. We also propose a learned threshold to facilitate the quantization process. 

\begin{figure*}
     \centering
     \begin{subfigure}[b]{0.24\textwidth}
         \centering
         \includegraphics[width=\textwidth]{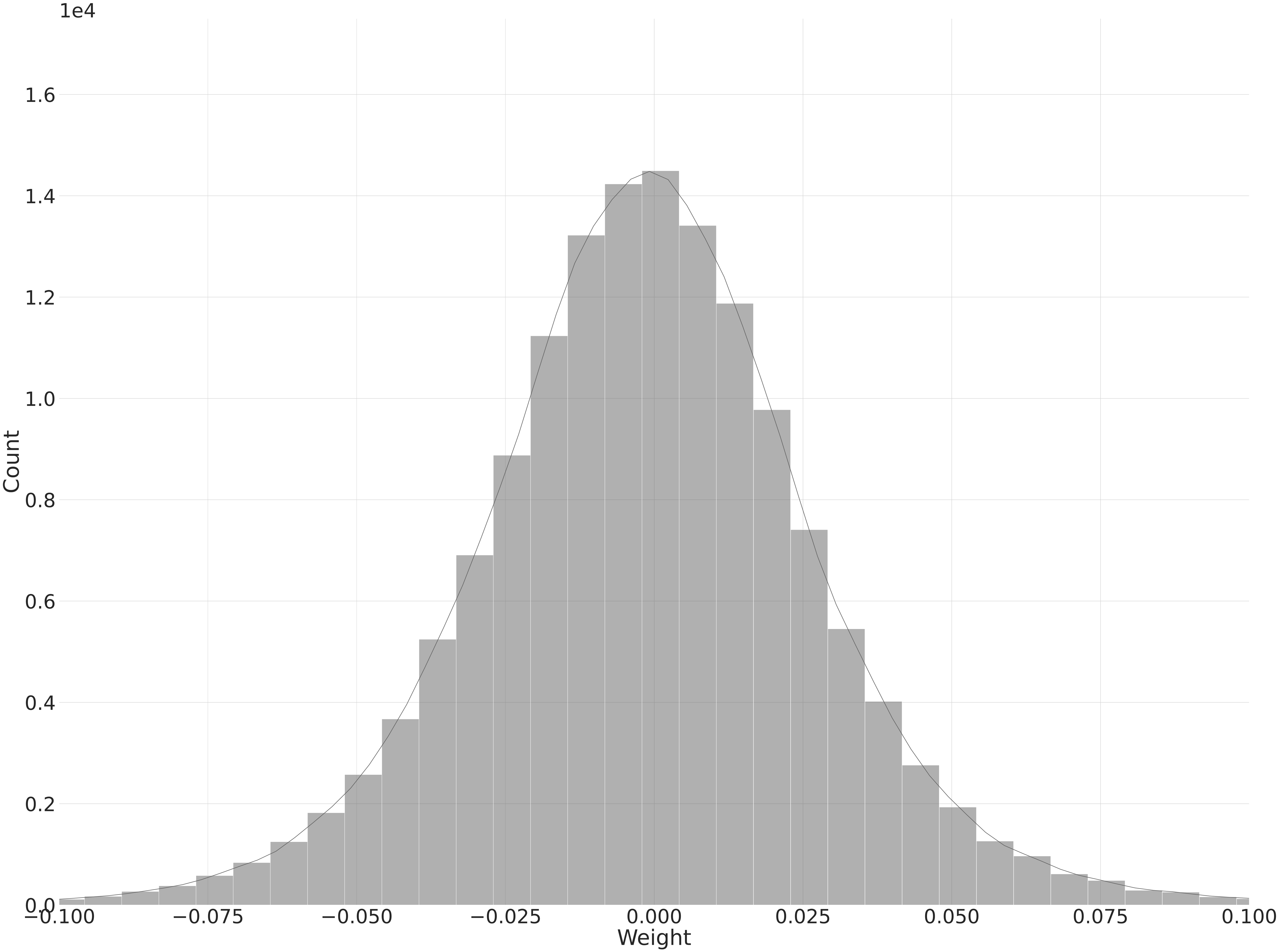}
         \caption{$t$=0}
         \label{fig:y equals x}
     \end{subfigure}
     \begin{subfigure}[b]{0.24\textwidth}
         \centering
         \includegraphics[width=\textwidth]{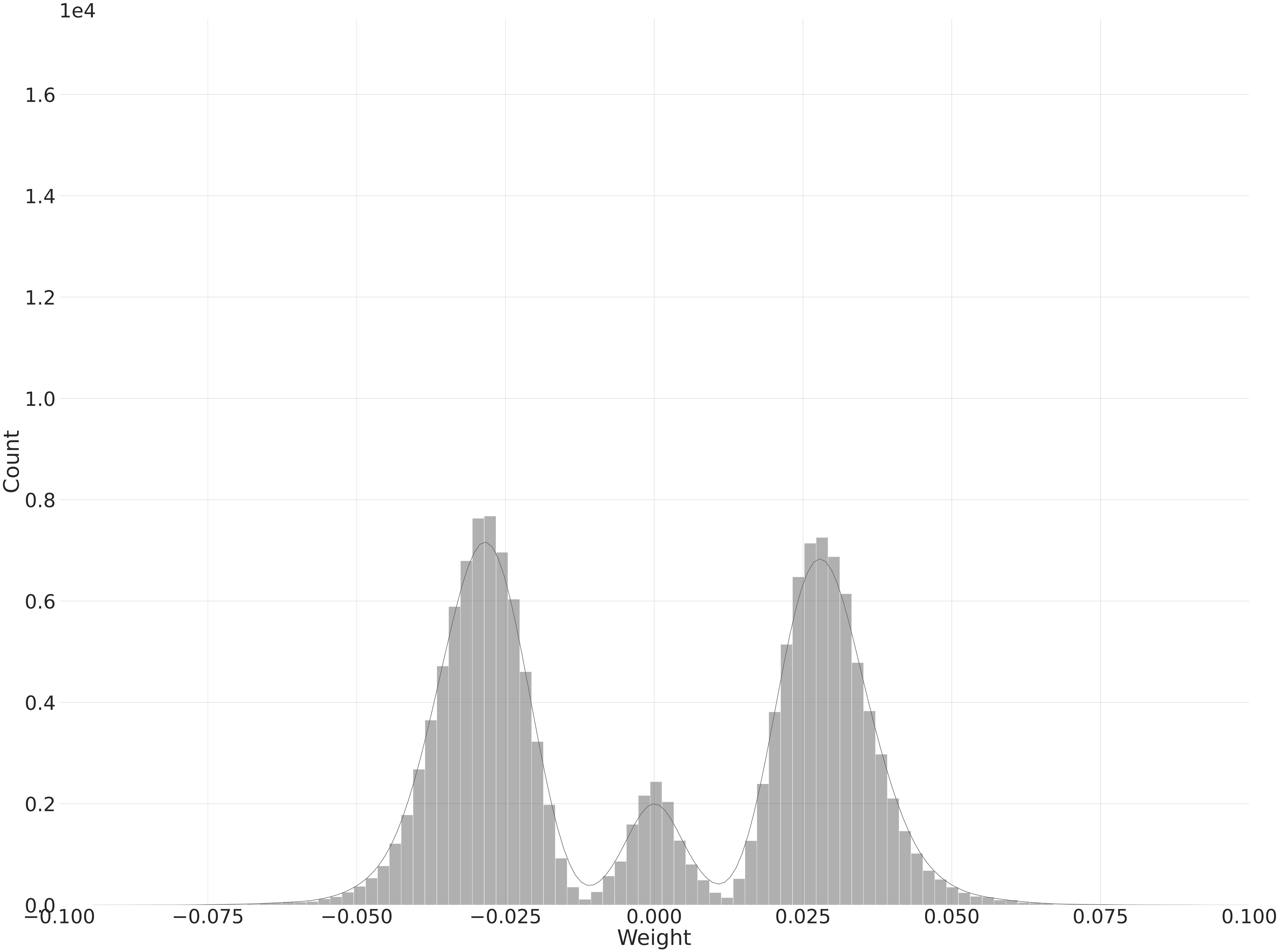}
         \caption{$t$=0.5}
         \label{fig:three sin x}
     \end{subfigure}
     \begin{subfigure}[b]{0.24\textwidth}
         \centering
         \includegraphics[width=\textwidth]{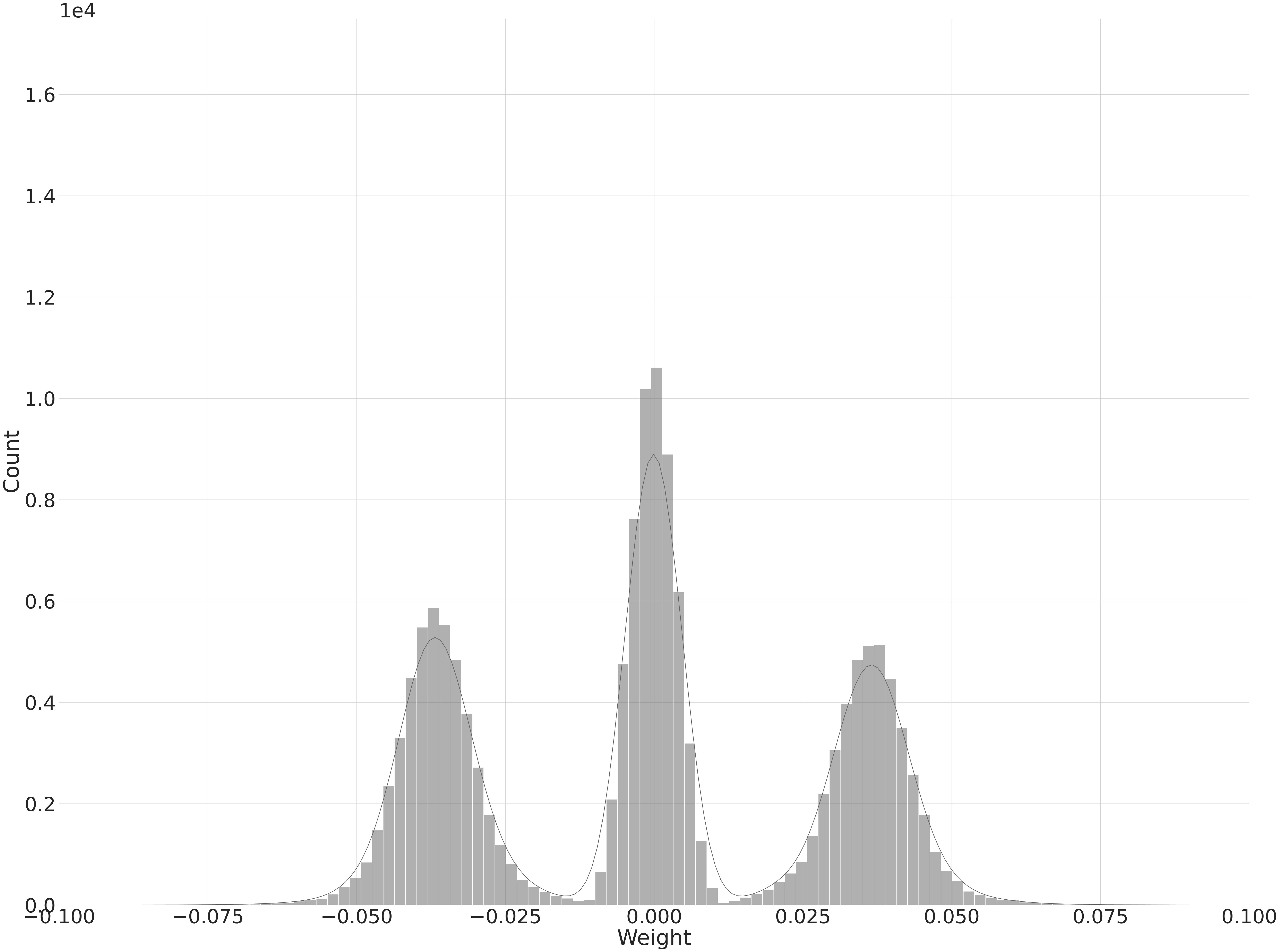}
         \caption{$t$=0.7}
         \label{fig:three sin x}
     \end{subfigure}
     \begin{subfigure}[b]{0.24\textwidth}
         \centering
         \includegraphics[width=\textwidth]{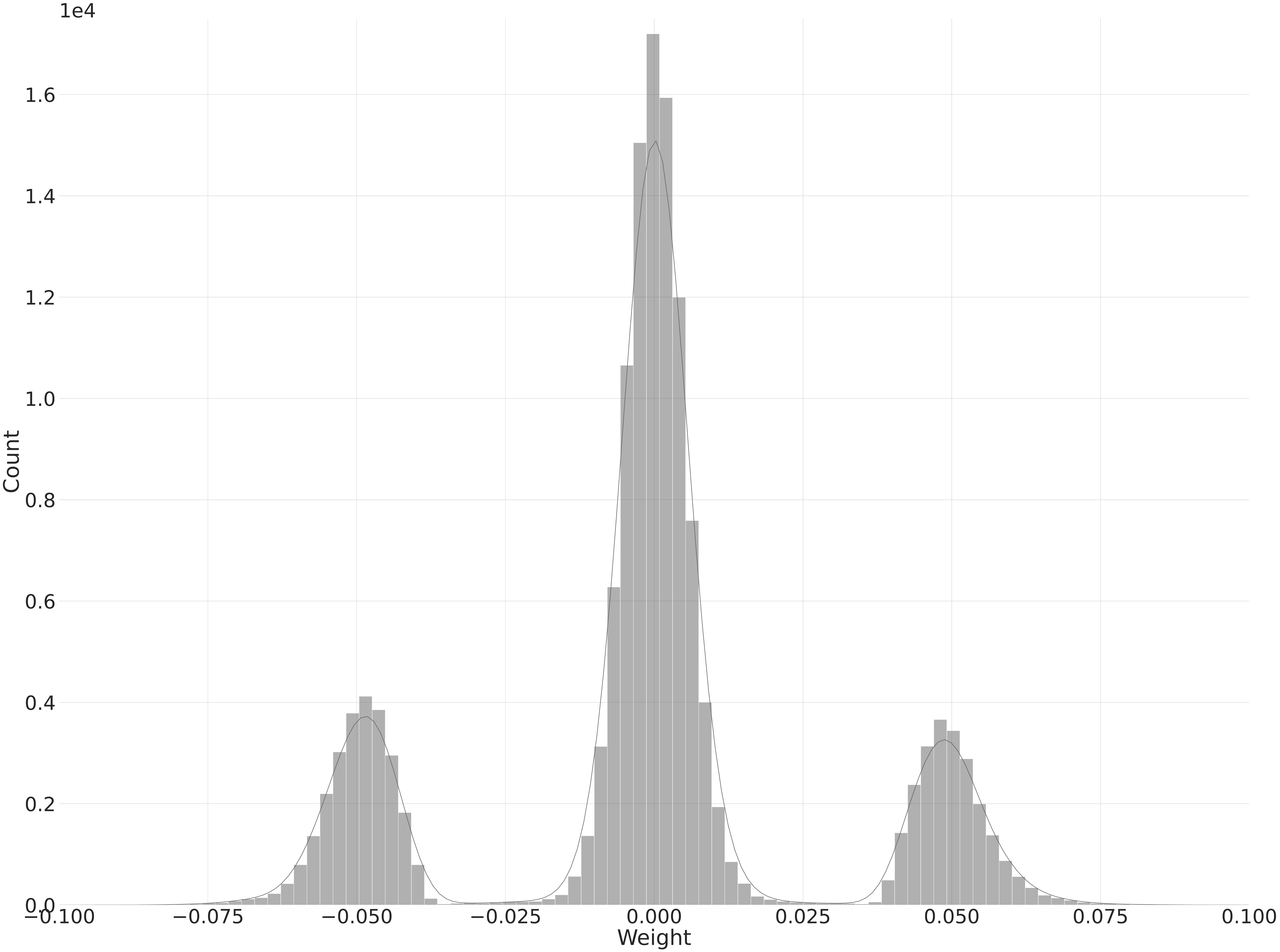}
         \caption{$t$=0.9}
         \label{fig:three sin x}
     \end{subfigure}
    \caption{\textcolor{black}{The weight distribution of a layer after applying the proposed regularization term $L_d$ (Eq.~\eqref{eq:loss}). (a) is a ResNet-18 baseline model. 
    Prior to ternary quantization, the regularisation term $L_{d}$ divides the full precision weight values into three clusters. As the sparsity factor $t$ increases, more weight values are moving closer to zero.}}
    \label{img:weight_dist}
\end{figure*}
\subsection{Pushing $\mathbf{W}$ close to $\mathbf{\hat{W}}$ before Quantization}

Given a regular objective function $L$, we formulate the optimization process as:
\begin{equation}
    \label{eq:loss}
     \min_{\mathbf{W}} J(\mathbf{W})= L(\mathbf{W})+L_{d}(\mathbf{W},\Delta).
\end{equation}
The regularization term $L_{d}$ is defined as: 
\begin{equation}
    \label{eq:ltr}
    L_{d}(\mathbf{W},\Delta)=\frac{1}{n}\left(\texttt{diag}(\mathbf{W}^\top\mathbf{\hat{W}}-\mathbf{I})\right)^2 
\end{equation}
\begin{equation*}
    s.t.~~\mathbf{\hat{W}}=\texttt{Ternary}(\mathbf{W},\Delta),
\end{equation*}
where $\mathbf{I}$ is the identity matrix, and $\texttt{diag}(\cdot)$ returns the diagonal elements of a matrix. The threshold $\Delta$ is:
\begin{equation}
\label{eq:prune}
   \Delta=\texttt{T}(t)=\texttt{ValueAtIndex}(\texttt{sort}(|\mathbf{W}|,\texttt{idx}), 
\end{equation}
\begin{equation*}
   s.t. ~~ \texttt{idx}=\lfloor t\times~m\times~n\rfloor, 0<t<1,
\end{equation*}
where $t$ controls the sparsity, i.e., percentage of zero values, of $\mathbf{\hat{W}}$ (Figure \ref{img:weight_dist}), and $\texttt{ValueAtIndex}([\cdot],\texttt{idx})$ returns a value of $[\cdot]$ at the index $\texttt{idx}$. During training, based on the work of \cite{liu2018rethinking}, we initially assign $t=0.5$, namely, removing $50\%$ values by their magnitude in $\mathbf{W}$ as the potential ternary references $\mathbf{\hat{W}}$, then $t$ is gradually increased to $t=0.7$, i.e., 70\% of $\mathbf{\hat{W}}$ is zero. We use the quadratic term to keep $L(\mathbf{W})$ and $L_{d}(\mathbf{W},\Delta)$ at the same scale. Since $J(\mathbf{W})$ consists of two parts and is optimized in continuous space, the distance between $\mathbf{W}$ and $\mathbf{\hat{W}}$ will gradually decrease without much change in model accuracy. In practice, we observe that applying $L_{d}(\mathbf{W},\Delta)$ barely reduces the model accuracy.

Since we are training with hyperspherical settings, i.e., $\|\mathbf{w}_j\|_2=\|\mathbf{\hat{w}}_j\|_2=1$, the diagonal elements of $\mathbf{W}^\top\mathbf{\hat{W}}$ in Eq.~\eqref{eq:ltr} denotes the cosine similarities between $\mathbf{w_j}$ and $\mathbf{\hat{w}_j}$. Minimizing $L_{d}$ is equivalent to pushing $\mathbf{w}_j$ close to $\mathbf{\hat{w}}_j$, namely, making part of the magnitude of weight values close to $\alpha$ (Eq.~\eqref{eq_ter}) while the rest close to zero (Figure \ref{img:weight_dist}).
In addition, the gradient of $L_{d}$ can adjust $\frac{\partial J}{\partial {\mathbf{W}}}$, which is similar to the works of \cite{hou2018loss_31,yang2019quantization_22,zhou2018explicit_26.5,dbouk2020dbq_20}. 

\subsection{Rescaling the Gradient During Quantization}
\label{rescaling}
\begin{figure}
     \centering
     \begin{subfigure}[b]{0.4\textwidth}
         \centering
         \includegraphics[width=\textwidth]{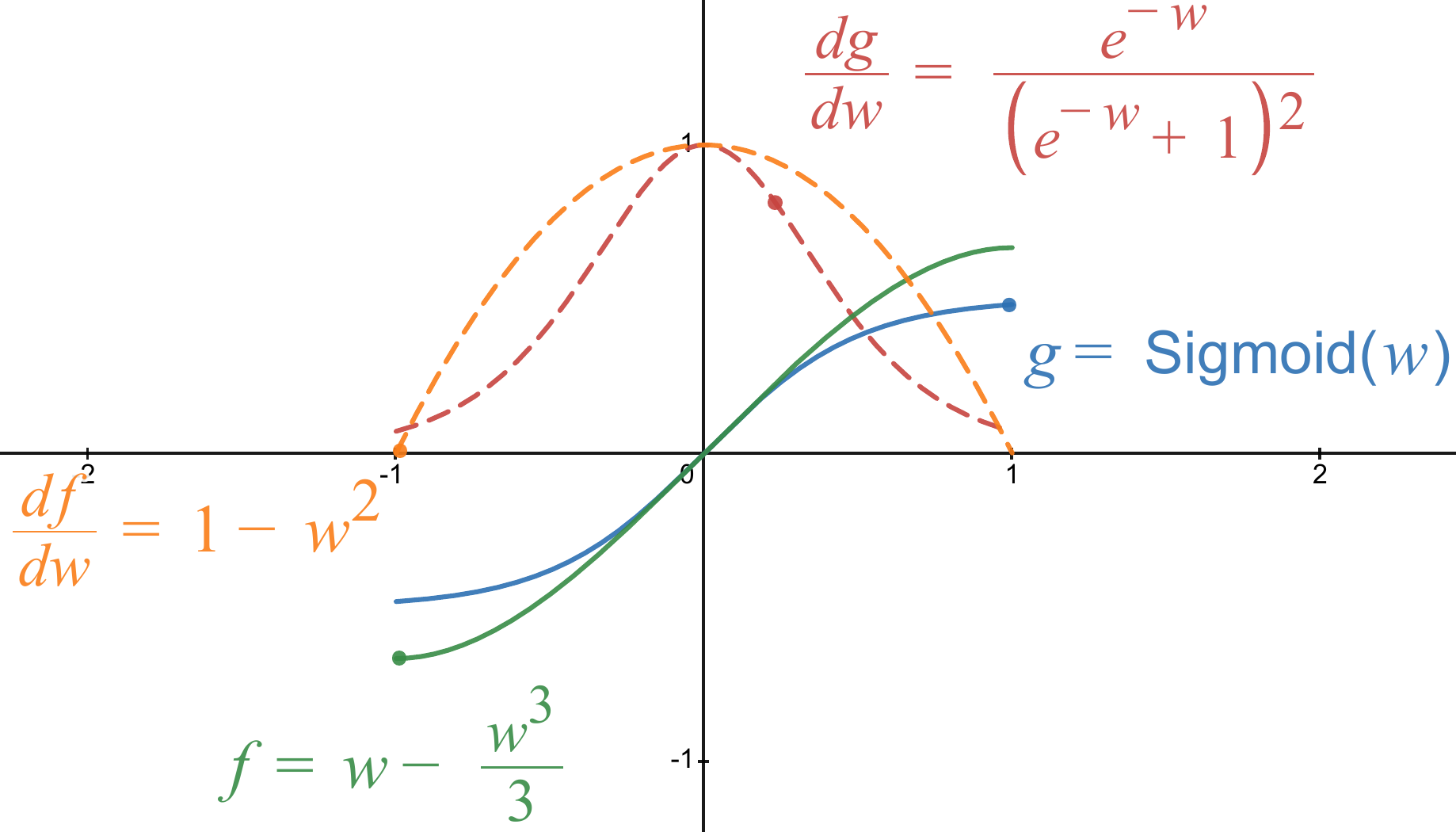}
         \label{fig:three sin x}
     \end{subfigure}
    \caption{Function $f$ and $g$ share the similar shape. $S=\frac{df}{dw}$ is an alteration of $\frac{dg}{dw}$ to re-scale the gradient and is much easier to calculate. }
    \label{img:scale}
\end{figure}
With $\mathbf{W}$ getting close to $\hat{\mathbf{W}}$ before the ternary quantization, we still need to convert the full precision weight to ternary values. We change the $\Delta$ to a learnable threshold $\bar\Delta$ and empirically initialize $\bar\Delta=\texttt{T}(t+0.1)$ via Eq.~\eqref{eq:prune}. The Eq.~\eqref{eq:loss} becomes:
\begin{equation}
\label{eq:lossdelta}
     \min_{\mathbf{W},\bar\Delta}J(\mathbf{W},\bar\Delta)= L(\mathbf{\hat{W}})+L_{d}(\mathbf{W},\bar\Delta).
\end{equation}
We introduce a gradient scaling factor $S$ and adopt STE \cite{bengio2013estimating} to bypass the non-differentiable
problem of $\texttt{Ternary}(\cdot)$:

\textbf{Forward:}
\begin{align}
\begin{split}
\hat{\mathbf{W}}={\texttt{Ternary}(\mathbf{W},\bar\Delta)};
\end{split}
\end{align}

\textbf{Backward:}
\begin{equation}
    \label{eq9}
    \frac{\partial J}{\partial {\mathbf{W}}}=\frac{\partial J }{\partial {\hat{\mathbf{W}}}} \frac{\partial \hat{\mathbf{W}} }{\partial {\mathbf{W}}}\underset{S \bar{T} E}{\approx} \frac{\partial J }{\partial {\hat{\mathbf{W}}}}\times S,
\end{equation}
where $S$ is defined as:
\begin{equation}
    S=\mathbf{1}-{\mathbf{W} \odot \mathbf{W}},
\end{equation}
and $\odot$ denotes the element-wise multiplication. Inspired by the work of \cite{yang2019quantization_22}, $1-w^2$ is used to re-scale the gradient of $w_{ij}$ as it shares similarities with the derivative of $\texttt{Sigmoid}(\cdot)$ (Figure \ref{img:scale}). $S$ is intuitive and easy to calculate. The gradient of $w_{ij}$ should be smaller when it gets farther away from $\bar\Delta$. 
This intuition is the same as the works of \cite{hou2018loss_31,zhou2018explicit_26.5,dbouk2020dbq_20}.

We simply apply the averaged gradients of $w_{ij}$ to update the quantization threshold $\bar\Delta$:
\begin{equation}
    \frac{\partial L}{\partial \bar\Delta} =\frac{1}{m\times n}\sum_{i=1}^{m}\sum_{j=1}^{n}{\frac{\partial L}{\partial w_{ij}}},\label{eq19}
\end{equation}
\begin{equation*}
    s.t.~~w_{ij}\in \mathbf{W}  \text{ and } w_{ij} \ne 0.
\end{equation*}
Figure \ref{img:weight_dist} shows that we can initialize $\bar\Delta$ with a smaller value and then gradually increase it until the near-zero portion is converted to zero during quantization. We should increase the threshold $\bar\Delta$ rapidly when the error is very large. As the training error becomes smaller, such increase should slow down.


\subsection{Implementation Details}

\subsubsection{Training Algorithm}
The proposed method is in Algorithm~\ref{alg:training}. We gradually increase the $t$ from $t_1=0.3$ to $t_2=0.7$ based on \cite{liu2018rethinking}. $t=0.7$ gives $\mathbf{\hat{W}}$ 70\% sparsity (Eq.~\eqref{eq:prune}). The overall process can be summarized as:
i) Fine-tuning from pre-trained model weights with hyperspherical learning architecture \cite{liu2017deephyperspherical}; ii) Initialize  $\bar\Delta=\texttt{T}(t_2+0.1)$ to convert near-zero weight values to zero; iii) Ternary quantization, updating the weights and $\bar\Delta$ through STE. Note that $\Delta$ is a global magnitude threshold \cite{blalock2020state} and $\bar\Delta$ is initialized globally and updated  by SGD in layer-wise manner. 
\begin{algorithm}
	\caption{HLA training approach}
	\label{alg:training}
	\begin{algorithmic}[1]
		\State \textbf{Input:} Input $\mathbf{x}$, a hyperspherical neural layer $\phi(\cdot)$, $t_1=0.3$ and $t_2=0.7$.
		\State \textbf{Result:} Quantized ternary networks for inference

        
        
        \State\textbf{1. Fine-tuning:}
        \State $t = t_1$
        \While{
		\textit{$t<t_2$}
		} 
	    
	    
	    \State $\Delta=\texttt{T}(t)$  \Comment{Update $\Delta$ }

		\While{
		\textit{not converged}
		} \label{lst:line:9}
		
		\State{
	    $y=\phi (\mathbf{W}^\top\mathbf{x})$ \Comment{Hyperspherical training}
	    }
	    \State{
            Minimize Eq.~\eqref{eq:loss}
	    }
	    \State{
	    Perform SGD, calculate $\frac{\partial J}{\partial {\mathbf{W}}}$, and update $\mathbf{W}$
	    }
		\EndWhile \label{lst:line:12}
		\State{$t+=0.04$} \Comment{Increase the $t$}
		\EndWhile

		\State\textbf{2. Ternary Quantization:}
		\State $\bar\Delta=\texttt{T}(t_2+0.1)$ 
		\While{\textit{not converged}} \label{lst:line:16}
		\State{
		$\mathbf{\hat{W}}=\texttt{Ternary}(\mathbf{W},\bar\Delta)$ }
	    \State{
	    $y=\phi ({\mathbf{\hat{W}}}^\top\mathbf{x})$
	    }
	    \State{Minimize Eq.~\eqref{eq:lossdelta}}
	    \State{
	    Get $\frac{\partial J}{\partial {\mathbf{{W}}}} $, $\frac{\partial J}{\partial {\bar\Delta}} $ via SGD; update $\mathbf{{W}}$, $\bar\Delta$ \Comment{Eq.~(\ref{eq9},\ref{eq19})}
	    }
		\EndWhile \label{lst:line:20}
	
	\end{algorithmic}
\end{algorithm}
\subsubsection{\textcolor{black}{Training Time}}
When training the ResNet-18 model with 8$\times$V100 and mixed-precision (16-bit), each epoch takes about 6 minutes. It takes about 100 epochs to obtain a hyperspherical ready-to-quantize model. The inner SGD loop (Line \ref{lst:line:9}-\ref{lst:line:12} in Algorithm \ref{alg:training}) takes about 10 epochs.
The ternary quantization loop (Line \ref{lst:line:16}-\ref{lst:line:20} in Algorithm \ref{alg:training}) takes about 200 epochs. 

\section{Discussion}
Our work shows that the model's angular information, \textit{i.e.}, the cosine similarity, connects sparsity and quantization on the hypersphere. We demonstrate how gradually adjusting the sparsity constraints can facilitate ternary quantization. 

Most ternary quantization works try to directly project full precision models into ternary ones. However, the discrepancy between the full precision and ternary values before quantization is more or less ignored. Only a few approaches \cite{ding2017three_40,hu2019cluster_24} attempt to minimize the discrepancy prior to quantization. Intuitively, the gradients during ternary quantization are inaccurate; therefore, it is inapproriate to use such gradients for distance optimization. Accurate gradients, which are produced by full precision weights, are better than estimated gradients in terms of reducing discrepancy. Our work shows that by making the full precision weights close to the ternary in the initial steps, we can improve the performance of ternary quantization. 

In mainstream ternary quantization works, such as TWN \cite{li2016twn} and TTQ \cite{zhu2016ttq}, due to the non-differentiable thresholds and unstable weight magnitudes, fixed thresholds and learned scaling factors are used to optimize ternary boundaries to determine which values should be zero. In our work, the clustered weight values with hyperspherical learning allows us to use the average gradient to update the quantization threshold $\bar\Delta$ directly, as the thresholding only needs a tiny increment to push the less important weights to zero.

Compared to temperature-based works \cite{yang2019quantization_22} using Sigmoid function for projection and layer-wise training, our proposed method is simple yet effective: by combining a bell-shaped gradient re-scaling factor $S$ with ternary quantization, we achieve a better efficiency-accuracy trade-off. In addition, our method does not need too many hyperparameters and sophisticated learned variables \cite{li2020rtn_23,yang2019quantization_22}.
Compared to other loss-aware ternary quantization works \cite{zhou2018explicit_26.5,hou2018loss_31} using estimated gradients to optimize the loss penalty term, our method minimizes the loss function by using full precision weights before quantization, which is more accurate and efficient.



\section{Experiments}
Our method is evaluated on image classification and object detection tasks with the ImageNet dataset \cite{ILSVRC15}. The ResNet-18/50 \cite{he2016deep} and MobileNetV2 \cite{sandler2018mobilenetv2} architectures are for image classification. The Mask R-CNN \cite{he2017mask} with ResNet-50 FPN \cite{wu2019detectron2} is for object detection. The object detection tasks are performed on the MS COCO \cite{lin2014mscoco} dataset. The pre-trained weights are provided by the PyTorch zoo and Detectron2 \cite{wu2019detectron2}. 

\subsection{Experiment Setup}
For image classification, the batch size is 256. The weight decay is 0.0001, and the momentum of stochastic gradient descent (SGD) is 0.9. We use the cosine annealing schedule with restarts \cite{loshchilov2016sgdr} to adjust the learning rates. The initial learning rate is 0.01. 
All of the experiments use 16-bit Automatic Mixed Precision (AMP) from PyTorch to accelerate the training process. The first convolutional layer and last fully-connected layer are skipped when quantizing ResNet models. For the MobileNetV2, and Mask R-CNN, we only skip the first convolutional layer.
\subsection{Image Classification}

\begin{table}[t]
    \footnotesize
        \centering
        \begin{adjustbox}{max width=0.5\textwidth}
        \begin{tabular}{lllcc}
        \toprule
        \textbf{Models} &
        \textbf{Methods}
        & 
        \textbf{Bits \tiny(W/A)}
        & 
        \textbf{Acc.}
        \\
        \midrule
        \multirow{4}{6em}{ResNet-18\\Acc.: 69.76}& 
         \textsc{\textbf{HLA (Ours)}} & \textbf{2/16}  & \textbf{68.6} &   \\
         & \textsc{TWN \cite{li2016twn}} & 2/32  & 61.8  \\
        & \textsc{TTQ \cite{zhu2016ttq}} & 2/32  & 66.6 & \\
        & \textsc{INQ \cite{zhou2017incremental}} & 2/32  & 66.0  \\
        & \textsc{ADMM \cite{leng2018extremelyadmm}} & 2/32  & 67.0  \\
        & \textsc{LQ-Net \cite{zhang2018lqnet}} & 2/32  & 68.0  \\
        & \textsc{QN* \cite{yang2019quantization_22}} & 2/32  & 69.1  \\
        & \textsc{AdaRound$^+$ \cite{nagel2020up_adaround}} & 2/32  & 55.9\\  & \textsc{BRECQ$^+$ \cite{li2021brecq}} & 2/32  & 66.3  \\
        & \textsc{RTN \cite{li2020rtn_23}} & 2/32  & 68.5  \\
        
        \midrule
        \multirow{6}{6em}{ResNet-50\\Acc.: 76.15}&  
         \textbf{HLA (Ours)} & \textbf{2/16}  & \textbf{75.2}  \\
         
        & \textsc{TWN\cite{li2016twn}} & 2/32& 72.5   \\
        & \textsc{LQ-Net \cite{zhang2018lqnet}} & 2/32  & 75.1 \\
        & \textsc{QN* \cite{yang2019quantization_22}} & 2/32  & 75.2 \\
        & \textsc{AdaRound$^+$ \cite{nagel2020up_adaround}} & 2/32  & 47.5\\  & \textsc{BRECQ$^+$ \cite{li2021brecq}} & 2/32  & 72.4  \\
        \midrule
        \multirow{3}{6em}{MobileNetV2\\Acc.: 71.88}
        & \textsc{\textbf{HLA (Ours)}} & 2/16  & \textbf{58.7}   \\
        & \textsc{AdaRound$^+$ \cite{nagel2020up_adaround}} & 2/32  & 32.5\\
        & \textsc{BRECQ$^+$ \cite{li2021brecq}} & 2/8  & 56.3  \\
        & \textsc{DC \cite{wang2019haq}} & 2/32  & 58.0  \\
        \bottomrule
        \end{tabular}
        \end{adjustbox}
    \caption{Ternary quantization results on the ImageNet. ``Bits (W/A)'' denotes the bit-width of weight and activation. ``*'' indicates the highly time-consuming layer-wise training. ``$+$'' indicates post-training quantization.} 
    \label{tab_w2a32}
\end{table} 

We compare our results with leading ternary quantization results from BRECQ~\cite{li2021brecq}, RTN~\cite{li2020rtn_23}, and AdaRound~\cite{nagel2020up_adaround}. We also compare our work with existing milestone methods, such as TWN~\cite{li2016twn}, TTQ \cite{zhu2016ttq}, Deep Compression (DC)~\cite{han2015deep}, INQ ~\cite{zhou2017incremental}, ADMM~\cite{leng2018extremelyadmm}, LQ-Net~\cite{zhang2018lqnet}, and Quantization Networks (QN)~\cite{yang2019quantization_22}. 

The image classification results (Table~\ref{tab_w2a32}) show that our proposed method outperforms most of the leading ternary quantization methods. The top-1 test accuracy of ResNet-18 is slightly lower than the work of QN~\cite{yang2019quantization_22}. However, our method does not need time-consuming layer-wise training. We have the same results on ResNet-50. The works of RTN \cite{li2020rtn_23} and QN \cite{yang2019quantization_22} have similar results as ours, however, assigning more than four learned parameters for each layer is inconvenient and complicated. 

For MobileNetV2, we quantize the last fully-connected layer as the work of \cite{han2015deep,li2021brecq} and achieve a good result. MobileNet is a lightweight model with fewer multiplications and additions. The works of \cite{dbouk2020dbq_mob,gope2020ternary_mob,sung2015resiliency_mob} demonstrate that the less complex model is vulnerable to ternary quantization. Quantizing the pointwise layer of MobileNet leads to significant accuracy loss \cite{gope2020ternary_mob}. That is why the fully ternary quantized methods \cite{han2015deep,li2021brecq} have a lower accuracy.

\subsection{Object Detection and Segmentation}
We test our method on the Mask R-CNN \cite{he2017mask} architecture with ResNet-50 \cite{he2016deep} backbone, similar to earlier work \cite{li2021brecq}, to verify its generalizability (Table \ref{tab_coco_quant}). As stated in Algorithm~\ref{alg:training}, we apply our approach to the original Mask R-CNN source code. The pre-trained model and the original source code are obtained from Detectron2~\cite{wu2019detectron2}. We replicate Detectron2's hyperparameter settings and training methods.

\begin{table}[t]
    \footnotesize
        \centering
        \begin{adjustbox}{max width=\textwidth}
        \begin{tabular}{cccccc}
        \toprule
         \textbf{Methods} & \textbf{ Bits}\tiny\textbf{ (W/A)} & \textbf{AP$^{bb}$}&\textbf{AP$^{mk}$} \\
        \midrule
        \textsc{Baseline} &32/32&{38.5}&{35.2}\\
        \midrule
         \textsc{\textbf{HLA (Ours)}} &2/16&{35.8}&{32.5}\\ 
         \textsc{BRECQ \cite{li2021brecq}} &2/8&34.2&-\\
        
        \bottomrule
        \end{tabular}
        \end{adjustbox}
        	\caption{The model size and Average Precision (AP) with bounding box (bb) and mask (mk) are compared. The \textsc{``Baseline''} is Mask-RCNN from Detectron2.
	}
        \label{tab_coco_quant}
    \end{table}   

\section{\textcolor{black}{Ablation Study}}
\label{abst}
In this section, we study the impact of the sparsity constraint $t$, the re-scaling factor $S$, and the hyperspherical training.
\subsection{\textcolor{black}{The Sparsity Constraint $t$ of $\mathbf{\hat{W}}$}}
The sparsity of $\mathbf{\hat{W}}$ is controlled by $t$ and is connected to the zero portion of ternary weight values. The work of \cite{zhu2016ttq} shows that the sparsity can affect ternary quantization. However it only studies on CIFAR-10 with a smaller ResNet-20 model. We compare different settings of sparsity constraint $t$,  including fixed and gradually increased value, to observe the final ternary quantizaion results. We also compare different initial $\bar\Delta$ (Table \ref{tr_acc_ab}).

\begin{figure}[h]
\centering
\includegraphics[width=0.8\columnwidth]{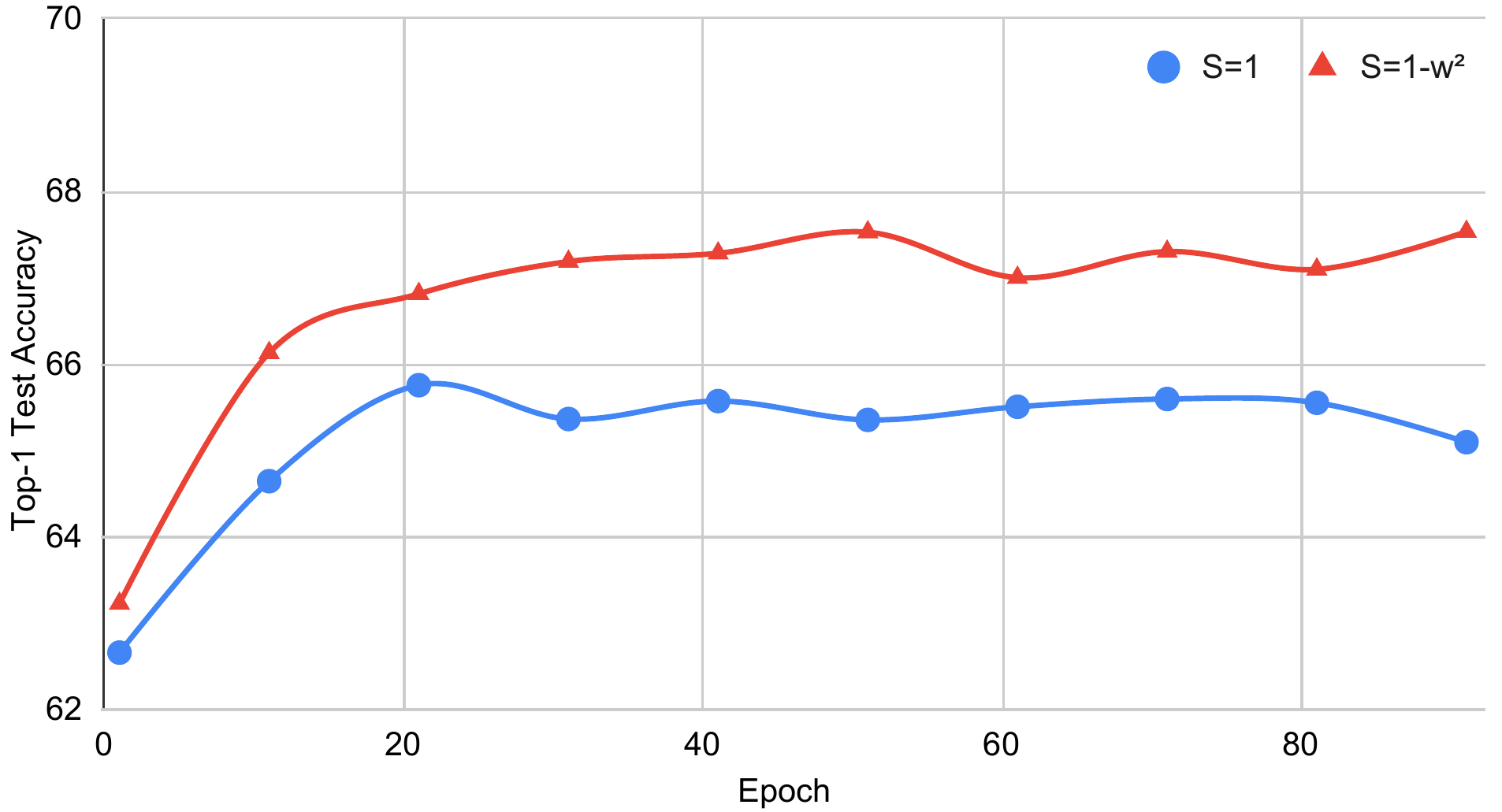} 
\caption{
The trend lines of top-1 test accuracy of ternary quantization of ResNet-18 on the ImageNet dataset with and without the re-scaling factor $S$.
}
\label{fig:sa_ab}
\end{figure}

\begin{figure}[h]
\centering
\includegraphics[width=0.8\columnwidth]{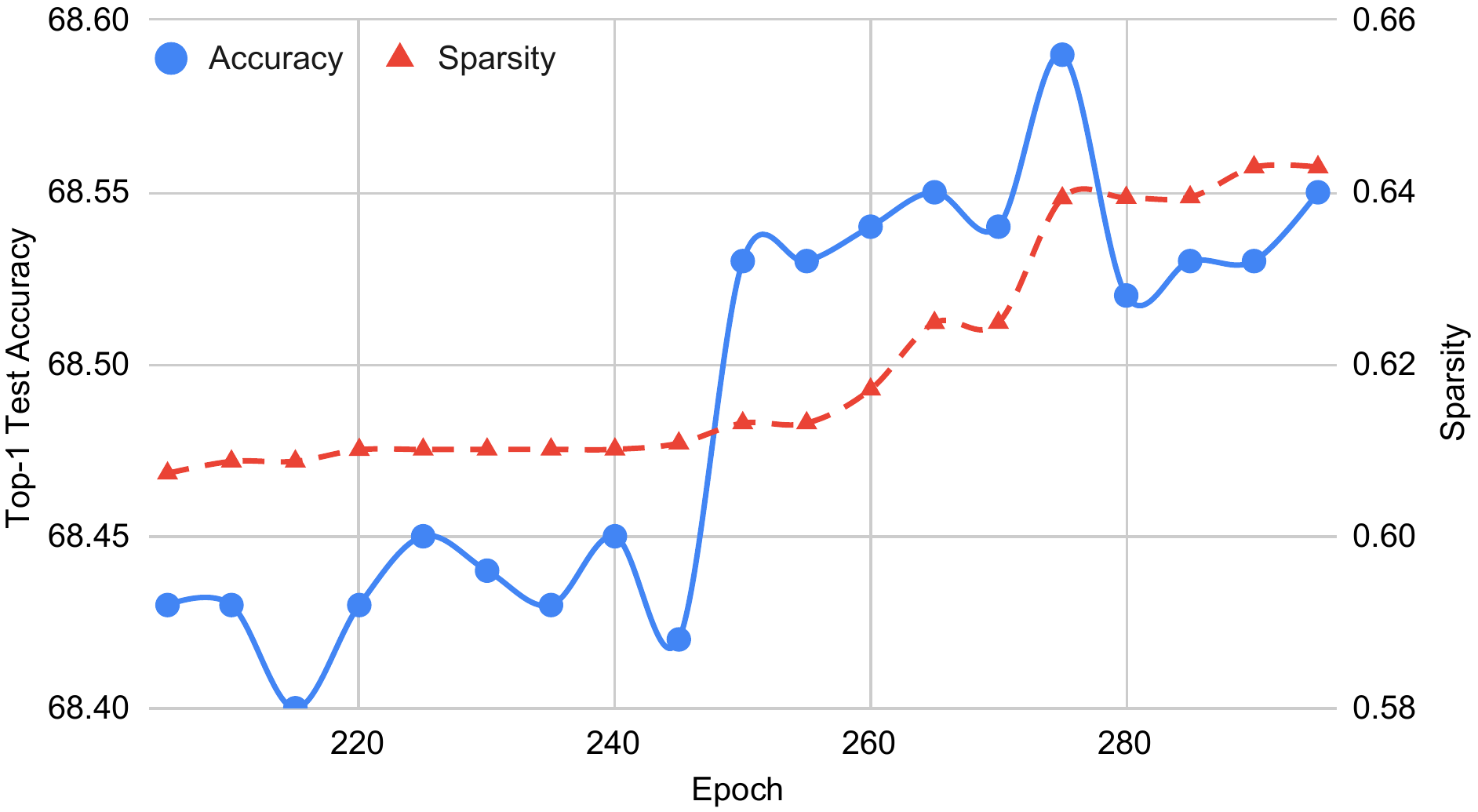} 
\caption{
The trend lines of top-1 test accuracy and sparsity of ternary quantization of ResNet-18 on the ImageNet dataset. Since the sparsity grows very slowly, we only plot the samples from 200 to 300 epochs. The red dashed line and right hand side axis indicate the model sparsity.
}
\label{fig:spacc_ab}
\end{figure}

\begin{table}[!ht]
      \footnotesize

        \centering

        \begin{adjustbox}{max width=\textwidth}
        \begin{tabular}{lcccc}
        \toprule
        
         ~ & $\Delta_{t=0.5}$ &  $\Delta_{t=0.7}$ &  $\Delta_{t=0.9}$ & $\Delta_{t=0.3\rightarrow0.7}$  \\ 
         \midrule
        $\bar\Delta_{t=0.7}$ & 63.00 & 65.57 & 66.85 & 66.94  \\ 
        $\bar\Delta_{t=0.65}$ & 66.52 & 66.38 & 67.59 & 67.69  \\ 
        $\bar\Delta_{t=0.6}$ & 65.10 & 66.15 & 66.86 & 68.03  \\ 
        
    \bottomrule
        \end{tabular}
        \end{adjustbox}
        	\caption{The impact of $t$ on $\Delta$ and $\bar\Delta$ to the top-1 accuracy of ternary quantization on ResNet-18 with the ImageNet dataset. We take the results of the $99$-th epoch considering the training cost.
	}
        \label{tr_acc_ab}
    \vspace{-2mm}
    \end{table}
We compare different combinations of the sparsity constraint $t$ and the initial ternary threshold $\bar\Delta$. A good way of fine-tuning is gradually increasing $t$ from $0.3$ to $0.7$, and take $\bar\Delta=\texttt{T}(0.65)$ as initial value to get the ternary model. Over-pruned settings ($\bar\Delta>\Delta$) reduce the quantization performance.   The sparsity settings are based on the work of \cite{zhu2016ttq,liu2018rethinking}. Re-scaling factor $S$ (Figure \ref{fig:sa_ab}) 
improves ternary quantization. Figure \ref{fig:spacc_ab}
 shows the changes of accuracy and model sparsity. Since $\bar\Delta$ is updated by the mean gradient of $\mathbf{W}$ (Eq.~\eqref{eq19}), the sparsity grows very slowly.
\subsection{The Impact of Hyperspherical Learning and $L_d$}
We study the change of the cosine similarity between $\mathbf{W}$ and $\mathbf{\hat{W}}$ before and after applying our proposed $L_d$ to hyperspherical training.
Table~\ref{tr_cos} shows that different values of sparsity constraint $t$ affect the cosine similarity between
$\mathbf{W}$ and $\mathbf{\hat{W}}$. The use of $L_d$ with a proper $t$ increases the cosine similarity, which could be an indicator of the final quantization results.

\begin{table}[t]
    \footnotesize
        \centering
        \begin{adjustbox}{max width=\textwidth}
        \begin{tabular}{ccccc}
        \toprule
         \textbf{Methods} & \textbf{$\mathcal{S}im$}  & \textbf{Accuracy} \\
        
        \midrule
        \textsc{Baseline} &0.73&{69.76}\\
        \textsc{Hypersphere} &0.73&{70.02}\\
        \midrule
        
         \textsc{Hyper+$L_d, \Delta_{t=0.5}$} &0.93&{69.91}\\ 
         \textsc{Hyper+$L_d, \Delta_{t=0.7}$} &0.95&69.81\\
         \textsc{Hyper+$L_d, \Delta_{t=0.9}$} &0.81&69.73 \\
         \textsc{Hyper+$L_d, \Delta_{t=0.95}$} &0.67&{68.86}\\ 

        
        \bottomrule
        \end{tabular}
        \end{adjustbox}
        	\caption{The comparison of the average cosine similarity $\mathcal{S}im$ between $\mathbf{W}$ and $\mathbf{\hat{W}}$ of ResNet-18 models on the ImageNet dataset before ternary quantization. The \textsc{``Baseline''} is the pre-trained ResNet-18. \textsc{``Hypersphere''} means training with hyperspherical settings.
	}
        \label{tr_cos}
    \end{table}

         
         

        
        

\section{Conclusion}
We propose an innovative method, Hyperspherical Loss-aware Quantization, to produce sparse ternary weights by pre-processing the full precision weights. A major contribution of our method is the use of sparsity constraints $t$ to enhance the ternary quantization. Our work further demonstrates that sparsity and ternary quantization are linked through angular information on the hypersphere. The proposed loss penalty term $L_d$ reduces the cosine distance between the full precision and ternary weight values. In this way, the ready-to-quantize model is very close to the targeting ternary model. The proposed gradient re-scaling factor $S$ simulates the derivative of the Sigmoid function, which adjusts the gradients in a simple yet efficient way. Future research may incorporate our method with mixed precision quantization and ternary activation quantization.

\bibliography{aaai23}

\begin{thebibliography}{50}
\providecommand{\natexlab}[1]{#1}

\bibitem[{Bai, Wang, and Liberty(2018)}]{bai2018proxquant}
Bai, Y.; Wang, Y.-X.; and Liberty, E. 2018.
\newblock Proxquant: Quantized neural networks via proximal operators.
\newblock \emph{arXiv preprint arXiv:1810.00861}.

\bibitem[{Bengio, L{\'e}onard, and Courville(2013)}]{bengio2013estimating}
Bengio, Y.; L{\'e}onard, N.; and Courville, A. 2013.
\newblock Estimating or propagating gradients through stochastic neurons for
  conditional computation.
\newblock \emph{arXiv preprint arXiv:1308.3432}.

\bibitem[{Blalock et~al.(2020)Blalock, Gonzalez~Ortiz, Frankle, and
  Guttag}]{blalock2020state}
Blalock, D.; Gonzalez~Ortiz, J.~J.; Frankle, J.; and Guttag, J. 2020.
\newblock What is the state of neural network pruning?
\newblock \emph{Proceedings of machine learning and systems}, 2: 129--146.

\bibitem[{Chen et~al.(2020)Chen, Liu, Yu, Kautz, Shrivastava, Garg, and
  Anandkumar}]{BeidiChen2020AngularVH}
Chen, B.; Liu, W.; Yu, Z.; Kautz, J.; Shrivastava, A.; Garg, A.; and
  Anandkumar, A. 2020.
\newblock Angular Visual Hardness.
\newblock In \emph{International Conference on Machine Learning}.

\bibitem[{Cho et~al.(2021)Cho, Vahid, Adya, and Rastegari}]{cho2021dkm}
Cho, M.; Vahid, K.~A.; Adya, S.; and Rastegari, M. 2021.
\newblock DKM: Differentiable K-Means Clustering Layer for Neural Network
  Compression.
\newblock \emph{arXiv preprint arXiv:2108.12659}.

\bibitem[{Chung, Ahn, and Bengio(2016)}]{chung2016hierarchical}
Chung, J.; Ahn, S.; and Bengio, Y. 2016.
\newblock Hierarchical multiscale recurrent neural networks.
\newblock \emph{arXiv preprint arXiv:1609.01704}.

\bibitem[{Courbariaux, Bengio, and David(2015)}]{courbariaux2015binaryconnect}
Courbariaux, M.; Bengio, Y.; and David, J.-P. 2015.
\newblock Binaryconnect: Training deep neural networks with binary weights
  during propagations.
\newblock In \emph{Advances in neural information processing systems},
  3123--3131.

\bibitem[{Davidson et~al.(2018)Davidson, Falorsi, Cao, Kipf, and
  Tomczak}]{TimRDavidson2018HypersphericalVA}
Davidson, T.~R.; Falorsi, L.; Cao, N.~D.; Kipf, T.; and Tomczak, J.~M. 2018.
\newblock Hyperspherical variational auto-encoders.
\newblock In \emph{Uncertainty in Artificial Intelligence}.

\bibitem[{Dbouk et~al.(2020{\natexlab{a}})Dbouk, Sanghvi, Mehendale, and
  Shanbhag}]{dbouk2020dbq_20}
Dbouk, H.; Sanghvi, H.; Mehendale, M.; and Shanbhag, N. 2020{\natexlab{a}}.
\newblock DBQ: A differentiable branch quantizer for lightweight deep neural
  networks.
\newblock In \emph{European Conference on Computer Vision}, 90--106. Springer.

\bibitem[{Dbouk et~al.(2020{\natexlab{b}})Dbouk, Sanghvi, Mehendale, and
  Shanbhag}]{dbouk2020dbq_mob}
Dbouk, H.; Sanghvi, H.; Mehendale, M.; and Shanbhag, N. 2020{\natexlab{b}}.
\newblock DBQ: A differentiable branch quantizer for lightweight deep neural
  networks.
\newblock In \emph{European Conference on Computer Vision}, 90--106. Springer.

\bibitem[{Deng et~al.(2019)Deng, Guo, Xue, and Zafeiriou}]{deng2019arcface}
Deng, J.; Guo, J.; Xue, N.; and Zafeiriou, S. 2019.
\newblock Arcface: Additive angular margin loss for deep face recognition.
\newblock In \emph{Proceedings of the IEEE Conference on Computer Vision and
  Pattern Recognition}, 4690--4699.

\bibitem[{Ding, Wu, and Wu(2017)}]{ding2017three_40}
Ding, J.; Wu, J.; and Wu, H. 2017.
\newblock Three-Means Ternary Quantization.
\newblock In \emph{International Conference on Neural Information Processing},
  235--245. Springer.

\bibitem[{Gholami et~al.(2021)Gholami, Kim, Dong, Yao, Mahoney, and
  Keutzer}]{gholami2021survey}
Gholami, A.; Kim, S.; Dong, Z.; Yao, Z.; Mahoney, M.~W.; and Keutzer, K. 2021.
\newblock A survey of quantization methods for efficient neural network
  inference.
\newblock \emph{arXiv preprint arXiv:2103.13630}.

\bibitem[{Gope et~al.(2020)Gope, Beu, Thakker, and
  Mattina}]{gope2020ternary_mob}
Gope, D.; Beu, J.; Thakker, U.; and Mattina, M. 2020.
\newblock Ternary mobilenets via per-layer hybrid filter banks.
\newblock In \emph{Proceedings of the IEEE/CVF Conference on Computer Vision
  and Pattern Recognition Workshops}, 708--709.

\bibitem[{Han, Mao, and Dally(2015)}]{han2015deep}
Han, S.; Mao, H.; and Dally, W.~J. 2015.
\newblock Deep compression: Compressing deep neural networks with pruning,
  trained quantization and huffman coding.
\newblock \emph{arXiv preprint arXiv:1510.00149}.

\bibitem[{He et~al.(2017)He, Gkioxari, Doll{\'a}r, and Girshick}]{he2017mask}
He, K.; Gkioxari, G.; Doll{\'a}r, P.; and Girshick, R. 2017.
\newblock Mask r-cnn.
\newblock In \emph{Proceedings of the IEEE international conference on computer
  vision}, 2961--2969.

\bibitem[{He et~al.(2016)He, Zhang, Ren, and Sun}]{he2016deep}
He, K.; Zhang, X.; Ren, S.; and Sun, J. 2016.
\newblock Deep residual learning for image recognition.
\newblock In \emph{Proceedings of the IEEE conference on computer vision and
  pattern recognition}, 770--778.

\bibitem[{Hou and Kwok(2018)}]{hou2018loss_31}
Hou, L.; and Kwok, J.~T. 2018.
\newblock Loss-aware weight quantization of deep networks.
\newblock \emph{arXiv preprint arXiv:1802.08635}.

\bibitem[{Hu et~al.(2019)Hu, Li, Long, Hu, Zhu, Wang, and
  Gu}]{hu2019cluster_24}
Hu, Y.; Li, J.; Long, X.; Hu, S.; Zhu, J.; Wang, X.; and Gu, Q. 2019.
\newblock Cluster regularized quantization for deep networks compression.
\newblock In \emph{2019 IEEE International Conference on Image Processing
  (ICIP)}, 914--918. IEEE.

\bibitem[{Hwang and Sung(2014)}]{hwang2014fixed}
Hwang, K.; and Sung, W. 2014.
\newblock Fixed-point feedforward deep neural network design using weights+ 1,
  0, and- 1.
\newblock In \emph{2014 IEEE Workshop on Signal Processing Systems (SiPS)},
  1--6. IEEE.

\bibitem[{Jang, Gu, and Poole(2016)}]{jang2016categorical}
Jang, E.; Gu, S.; and Poole, B. 2016.
\newblock Categorical reparameterization with gumbel-softmax.
\newblock \emph{arXiv preprint arXiv:1611.01144}.

\bibitem[{Leng et~al.(2018)Leng, Dou, Li, Zhu, and Jin}]{leng2018extremelyadmm}
Leng, C.; Dou, Z.; Li, H.; Zhu, S.; and Jin, R. 2018.
\newblock Extremely low bit neural network: Squeeze the last bit out with admm.
\newblock In \emph{Proceedings of the AAAI Conference on Artificial
  Intelligence}, volume~32.

\bibitem[{Li, Zhang, and Liu(2016)}]{li2016twn}
Li, F.; Zhang, B.; and Liu, B. 2016.
\newblock Ternary weight networks.
\newblock \emph{arXiv preprint arXiv:1605.04711}.

\bibitem[{Li et~al.(2020)Li, Dong, Zhang, Bai, Chen, and Wang}]{li2020rtn_23}
Li, Y.; Dong, X.; Zhang, S.~Q.; Bai, H.; Chen, Y.; and Wang, W. 2020.
\newblock Rtn: Reparameterized ternary network.
\newblock In \emph{Proceedings of the AAAI Conference on Artificial
  Intelligence}, volume~34, 4780--4787.

\bibitem[{Li et~al.(2021)Li, Gong, Tan, Yang, Hu, Zhang, Yu, Wang, and
  Gu}]{li2021brecq}
Li, Y.; Gong, R.; Tan, X.; Yang, Y.; Hu, P.; Zhang, Q.; Yu, F.; Wang, W.; and
  Gu, S. 2021.
\newblock Brecq: Pushing the limit of post-training quantization by block
  reconstruction.
\newblock \emph{arXiv preprint arXiv:2102.05426}.

\bibitem[{Lin et~al.(2014)Lin, Maire, Belongie, Hays, Perona, Ramanan,
  Doll{\'a}r, and Zitnick}]{lin2014mscoco}
Lin, T.-Y.; Maire, M.; Belongie, S.; Hays, J.; Perona, P.; Ramanan, D.;
  Doll{\'a}r, P.; and Zitnick, C.~L. 2014.
\newblock Microsoft coco: Common objects in context.
\newblock In \emph{European conference on computer vision}, 740--755. Springer.

\bibitem[{Liu et~al.(2017{\natexlab{a}})Liu, Wen, Yu, Li, Raj, and
  Song}]{liu2017sphereface}
Liu, W.; Wen, Y.; Yu, Z.; Li, M.; Raj, B.; and Song, L. 2017{\natexlab{a}}.
\newblock Sphereface: Deep hypersphere embedding for face recognition.
\newblock In \emph{Proceedings of the IEEE conference on computer vision and
  pattern recognition}, 212--220.

\bibitem[{Liu et~al.(2016)Liu, Wen, Yu, and Yang}]{liu2016large}
Liu, W.; Wen, Y.; Yu, Z.; and Yang, M. 2016.
\newblock Large-margin softmax loss for convolutional neural networks.
\newblock In \emph{ICML}, volume~2, 7.

\bibitem[{Liu et~al.(2017{\natexlab{b}})Liu, Zhang, Li, Yu, Dai, Zhao, and
  Song}]{liu2017deephyperspherical}
Liu, W.; Zhang, Y.-M.; Li, X.; Yu, Z.; Dai, B.; Zhao, T.; and Song, L.
  2017{\natexlab{b}}.
\newblock Deep hyperspherical learning.
\newblock \emph{Advances in neural information processing systems}, 30.

\bibitem[{Liu et~al.(2018)Liu, Sun, Zhou, Huang, and
  Darrell}]{liu2018rethinking}
Liu, Z.; Sun, M.; Zhou, T.; Huang, G.; and Darrell, T. 2018.
\newblock Rethinking the value of network pruning.
\newblock \emph{arXiv preprint arXiv:1810.05270}.

\bibitem[{Loshchilov and Hutter(2016)}]{loshchilov2016sgdr}
Loshchilov, I.; and Hutter, F. 2016.
\newblock Sgdr: Stochastic gradient descent with warm restarts.
\newblock \emph{arXiv preprint arXiv:1608.03983}.

\bibitem[{Louizos et~al.(2018)Louizos, Reisser, Blankevoort, Gavves, and
  Welling}]{louizos2018relaxed}
Louizos, C.; Reisser, M.; Blankevoort, T.; Gavves, E.; and Welling, M. 2018.
\newblock Relaxed quantization for discretized neural networks.
\newblock \emph{arXiv preprint arXiv:1810.01875}.

\bibitem[{Martinez et~al.(2021)Martinez, Shewakramani, Liu, B{\^a}rsan, Zeng,
  and Urtasun}]{martinez_2020_pqf}
Martinez, J.; Shewakramani, J.; Liu, T.~W.; B{\^a}rsan, I.~A.; Zeng, W.; and
  Urtasun, R. 2021.
\newblock Permute, Quantize, and Fine-tune: Efficient Compression of Neural
  Networks.
\newblock \emph{arXiv preprint arXiv:2010.15703}.

\bibitem[{Nagel et~al.(2020)Nagel, Amjad, Van~Baalen, Louizos, and
  Blankevoort}]{nagel2020up_adaround}
Nagel, M.; Amjad, R.~A.; Van~Baalen, M.; Louizos, C.; and Blankevoort, T. 2020.
\newblock Up or down? adaptive rounding for post-training quantization.
\newblock In \emph{International Conference on Machine Learning}, 7197--7206.
  PMLR.

\bibitem[{Parikh, Boyd et~al.(2014)}]{parikh2014proximal}
Parikh, N.; Boyd, S.; et~al. 2014.
\newblock Proximal algorithms.
\newblock \emph{Foundations and trends{\textregistered} in Optimization}, 1(3):
  127--239.

\bibitem[{Park and Kwon(2019)}]{SungWooPark2019SphereGA}
Park, S.~W.; and Kwon, J. 2019.
\newblock Sphere Generative Adversarial Network Based on Geometric Moment
  Matching.
\newblock In \emph{Computer Vision and Pattern Recognition}.

\bibitem[{Rastegari et~al.(2016)Rastegari, Ordonez, Redmon, and
  Farhadi}]{rastegari2016xnor}
Rastegari, M.; Ordonez, V.; Redmon, J.; and Farhadi, A. 2016.
\newblock Xnor-net: Imagenet classification using binary convolutional neural
  networks.
\newblock In \emph{European conference on computer vision}, 525--542. Springer.

\bibitem[{Russakovsky et~al.(2015)Russakovsky, Deng, Su, Krause, Satheesh, Ma,
  Huang, Karpathy, Khosla, Bernstein, Berg, and Fei-Fei}]{ILSVRC15}
Russakovsky, O.; Deng, J.; Su, H.; Krause, J.; Satheesh, S.; Ma, S.; Huang, Z.;
  Karpathy, A.; Khosla, A.; Bernstein, M.; Berg, A.~C.; and Fei-Fei, L. 2015.
\newblock {ImageNet Large Scale Visual Recognition Challenge}.
\newblock \emph{International Journal of Computer Vision (IJCV)}, 115(3):
  211--252.

\bibitem[{Sandler et~al.(2018)Sandler, Howard, Zhu, Zhmoginov, and
  Chen}]{sandler2018mobilenetv2}
Sandler, M.; Howard, A.; Zhu, M.; Zhmoginov, A.; and Chen, L.-C. 2018.
\newblock Mobilenetv2: Inverted residuals and linear bottlenecks.
\newblock In \emph{Proceedings of the IEEE conference on computer vision and
  pattern recognition}, 4510--4520.

\bibitem[{Stock et~al.(2020)Stock, Joulin, Gribonval, Graham, and
  J{\'e}gou}]{stock2019killthebits}
Stock, P.; Joulin, A.; Gribonval, R.; Graham, B.; and J{\'e}gou, H. 2020.
\newblock And the bit goes down: Revisiting the quantization of neural
  networks.
\newblock In \emph{International Conference on Learning Representations
  (ICLR)}.

\bibitem[{Sung, Shin, and Hwang(2015)}]{sung2015resiliency_mob}
Sung, W.; Shin, S.; and Hwang, K. 2015.
\newblock Resiliency of deep neural networks under quantization.
\newblock \emph{arXiv preprint arXiv:1511.06488}.

\bibitem[{Wang et~al.(2019)Wang, Liu, Lin, Lin, and Han}]{wang2019haq}
Wang, K.; Liu, Z.; Lin, Y.; Lin, J.; and Han, S. 2019.
\newblock Haq: Hardware-aware automated quantization with mixed precision.
\newblock In \emph{Proceedings of the IEEE conference on computer vision and
  pattern recognition}, 8612--8620.

\bibitem[{Wang et~al.(2018)Wang, Hu, Zhang, Zhang, Liu, and
  Cheng}]{wang2018two_27}
Wang, P.; Hu, Q.; Zhang, Y.; Zhang, C.; Liu, Y.; and Cheng, J. 2018.
\newblock Two-step quantization for low-bit neural networks.
\newblock In \emph{Proceedings of the IEEE Conference on computer vision and
  pattern recognition}, 4376--4384.

\bibitem[{Wu et~al.(2019)Wu, Kirillov, Massa, Lo, and
  Girshick}]{wu2019detectron2}
Wu, Y.; Kirillov, A.; Massa, F.; Lo, W.-Y.; and Girshick, R. 2019.
\newblock Detectron2.
\newblock \url{https://github.com/facebookresearch/detectron2}.

\bibitem[{Yang et~al.(2019)Yang, Shen, Xing, Tian, Li, Deng, Huang, and
  Hua}]{yang2019quantization_22}
Yang, J.; Shen, X.; Xing, J.; Tian, X.; Li, H.; Deng, B.; Huang, J.; and Hua,
  X.-s. 2019.
\newblock Quantization networks.
\newblock In \emph{Proceedings of the IEEE/CVF Conference on Computer Vision
  and Pattern Recognition}, 7308--7316.

\bibitem[{Yin et~al.(2019)Yin, Lyu, Zhang, Osher, Qi, and
  Xin}]{yin2019understanding}
Yin, P.; Lyu, J.; Zhang, S.; Osher, S.; Qi, Y.; and Xin, J. 2019.
\newblock Understanding straight-through estimator in training activation
  quantized neural nets.
\newblock \emph{arXiv preprint arXiv:1903.05662}.

\bibitem[{Zhang et~al.(2018)Zhang, Yang, Ye, and Hua}]{zhang2018lqnet}
Zhang, D.; Yang, J.; Ye, D.; and Hua, G. 2018.
\newblock Lq-nets: Learned quantization for highly accurate and compact deep
  neural networks.
\newblock In \emph{Proceedings of the European conference on computer vision
  (ECCV)}, 365--382.

\bibitem[{Zhou et~al.(2017)Zhou, Yao, Guo, Xu, and Chen}]{zhou2017incremental}
Zhou, A.; Yao, A.; Guo, Y.; Xu, L.; and Chen, Y. 2017.
\newblock Incremental network quantization: Towards lossless cnns with
  low-precision weights.
\newblock \emph{arXiv preprint arXiv:1702.03044}.

\bibitem[{Zhou et~al.(2018)Zhou, Yao, Wang, and Chen}]{zhou2018explicit_26.5}
Zhou, A.; Yao, A.; Wang, K.; and Chen, Y. 2018.
\newblock Explicit loss-error-aware quantization for low-bit deep neural
  networks.
\newblock In \emph{Proceedings of the IEEE conference on computer vision and
  pattern recognition}, 9426--9435.

\bibitem[{Zhu et~al.(2016)Zhu, Han, Mao, and Dally}]{zhu2016ttq}
Zhu, C.; Han, S.; Mao, H.; and Dally, W.~J. 2016.
\newblock Trained ternary quantization.
\newblock \emph{arXiv preprint arXiv:1612.01064}.

\end{thebibliography}




\end{document}